\documentclass[sigconf,authorversion,nonacm]{acmart}
\AtBeginDocument{%
  }

\setcopyright{none}
\renewcommand\footnotetextcopyrightpermission[1]{}
\usepackage{cleveref}
\usepackage{subcaption}
\usepackage{multirow}

\newcommand{\ie}{\textit{i}.\textit{e}.}
\newcommand{\eg}{\textit{e}.\textit{g}.}

\begin{document}

\title{Hierarchic-EEG2Text: Assessing EEG-To-Text Decoding across Hierarchical Abstraction Levels}

\author{Anupam Sharma}
\email{sharmaanupam@iitgn.ac.in}
\affiliation{%
  \institution{IIT Gandhinagar}
  \city{Gandhinagar}
  \state{Gujarat}
  \country{India}
}

\author{Harish Katti}
\email{kattih2@nih.gov}
\affiliation{%
  \institution{NIMH, NIH}
  \country{USA}
}

\author{Prajwal Singh}
\email{singh\_prajwal@iitgn.ac.in}
\affiliation{%
  \institution{IIT Gandhinagar}
  \city{Gandhinagar}
  \state{Gujarat}
  \country{India}
}

\author{Shanmuganathan Raman}
\email{shanmuga@iitgn.ac.in}
\affiliation{%
  \institution{IIT Gandhinagar}
  \city{Gandhinagar}
  \state{Gujarat}
  \country{India}
}

\author{Krishna Miyapuram}
\email{kprasad@iitgn.ac.in}
\affiliation{%
  \institution{IIT Gandhinagar}
  \city{Gandhinagar}
  \state{Gujarat}
  \country{India}
}


\begin{abstract}

An electroencephalogram (EEG) records the spatially averaged electrical activity of neurons in the brain, measured from the human scalp. Prior studies have explored EEG-based classification of objects or concepts, often for passive viewing of briefly presented image or video stimuli, with limited classes. Because EEG exhibits a low signal-to-noise ratio, recognizing fine-grained representations across a large number of classes remains challenging; however, abstract-level object representations may exist. In this work, we investigate whether EEG captures object representations across multiple hierarchical levels, and propose episodic analysis, in which a Machine Learning (ML) model is evaluated across various, yet related, classification tasks (episodes). Unlike prior episodic EEG studies that rely on fixed or randomly sampled classes of equal cardinality, we adopt hierarchy-aware episode sampling using \textit{WordNet} to generate episodes with variable classes of diverse hierarchy. We also present the largest episodic framework in the EEG domain for detecting observed text from EEG signals in the PEERS dataset, comprising $931538$ EEG samples under $1610$ object labels, acquired from $264$ human participants (subjects) performing controlled cognitive tasks, enabling the study of neural dynamics underlying perception, decision-making, and performance monitoring.

We examine how the semantic abstraction level affects classification performance across multiple learning techniques and architectures, providing a comprehensive analysis. The models tend to improve performance when the classification categories are drawn from higher levels of the hierarchy, suggesting sensitivity to abstraction. Our work highlights abstraction depth as an underexplored dimension of EEG decoding and motivates future research in this direction.

\end{abstract}

\ccsdesc[500]{Applied computing~Bioinformatics}
\ccsdesc[500]{Computing methodologies~Multi-task learning}

\keywords{EEG2Text, BCI, Hierarchical Modeling, Meta-Learning}


\settopmatter{printfolios=true}
\maketitle

\section{Introduction}
\label{sec:introduction}

Electroencephalography (EEG) is a non-invasive and cost-effective technique for measuring electrical activity in the human brain. EEG records spatially averaged electrical activity across the dendrites of neurons in the underlying brain tissue, measured from the human scalp via electrodes~\cite{cohen2014analyzing}. Due to spatial averaging, EEG does not capture all underlying neural activity, limiting the amount of fine-grained information that can be recovered from the signal~\cite{cohen2014analyzing}.

Despite these limitations, several studies have attempted to develop machine learning (ML) models that map EEG signals to the corresponding stimuli. Often, the stimuli are visual, and EEG is recorded while humans passively view images from a fixed set of categories. In these settings, ML models are trained to classify EEG signals into the corresponding category~\cite{spampinato2017deep, tirupattur2018thoughtviz, kaneshiro2015representational, singh2024learning, singh2023eeg2image, zhu2024eeg}, where the number of categories typically ranges from $10$ to $80$. In other studies, EEG signals are mapped to visually presented words or text~\cite{liu2024eeg2text, tao2025see, hollenstein2018zuco, hollenstein2020zuco}, where the number of categories equals the vocabulary size, which can be in the thousands. For example, the Zuco dataset~\cite{hollenstein2018zuco} contains approximately $21$K words.

While these studies demonstrate that EEG contains information related to the presented stimulus, our findings show that decoding performance degrades as the number of categories increases. This raises a fundamental question: rather than asking if EEG can support fine-grained classification over a large label space, we ask how much information it captures at varying levels of abstraction. Prior work in cognitive neuroscience has shown that neural representations of objects during visual processing evolve hierarchically over time~\cite {cichy2014resolving,carlson2013representational}. However, most EEG decoding studies evaluate performance using a flat label space, without explicitly considering the semantic hierarchy among categories. We further note that many existing datasets rely on abrupt presentation of visual stimuli, such as images or videos, during passive viewing tasks. Such stimuli are likely to elicit stereotyped brain responses across large regions~\cite{yamins2014performance}, which may benefit classification models but may not reflect real-world perception, decision-making, or performance monitoring. In contrast, tasks involving real-world object concepts are likely to introduce greater subjectivity across participants. In this work, we investigate whether EEG captures abstract, hierarchical representations of objects during such challenging real-world tasks. Instead of evaluating models jointly across all categories, we propose an episodic evaluation framework in which each episode consists of a smaller classification task. Unlike most episodic learning studies in the EEG domain, which construct episodes from random samples, we design episodes that follow a semantic hierarchy. This allows us to probe EEG representations at multiple levels of abstraction.

To this end, we leverage the Penn Electrophysiology of Encoding and Retrieval Study (PEERS) dataset~\cite{ds004395:2.0.0}, which consists of EEG signals recorded while participants viewed approximately $1.6$K words representing real-world objects. The large vocabulary and number of samples make this dataset suitable for episodic analysis. To construct the semantic hierarchy, we use NLTK's WordNet~\cite{bird2006nltk, princetonuniversity_2019_wordnet}, which links words and concepts via the \textit{IS-A} relationship. WordNet groups synonyms into synsets, and synsets are connected via \textit{IS-A} relations, which we use to build a Directed Acyclic Graph (DAG). The leaf nodes correspond to words in the PEERS dataset, while internal nodes represent synsets. Each internal node defines a classification task over its descendant leaf nodes, and sampling nodes at different depths of the DAG yields episodes at varying levels.

By using this strategy, we construct the largest episodic EEG-to-text decoding framework, comprising $931538$ EEG samples under $1610$ labels from $264$ subjects. We evaluate both non-episodic and episodic approaches, including meta-learning algorithms, across multiple modern architectures. Our results show that while modern architectures (even with self-supervised pretraining) fail under non-episodic evaluation, they perform comparatively better, though near chance, under hierarchical episodic evaluation. Moreover, performance improves as fine-grained classes are replaced with more abstract classes. These findings suggest that while EEG may not capture fine-grained representations, it does encode information at higher levels of semantic abstraction. Our contribution is a controlled evaluation framework that systematically examines which levels of semantic abstraction are accessible from the EEG during realistic cognitive tasks.
\section{Background}
\label{sec:background}

This section briefly introduces the evaluation and learning approaches used in our work.

\subsection{Episodic Vs Non-Episodic Evaluation}

\subsubsection{Non-Episodic Evaluation}
\label{subsubsec:non-episodic-eval}
In this work, we refer to the evaluation of an ML model in a typical classification task as non-episodic evaluation. Assume a Deep Neural Network (DNN) model trained for classifying samples into a set of classes $C_{train}$. When the model is evaluated on the task, the set of classes in the test set $C_{eval}$ typically remains the same as $C_{train}$, and the number of units in the network's last layer doesn't change during testing.

\subsubsection{Episodic Evaluation} Assume a DNN model trained for classifying samples into a set of classes $C_{train}$ is tested on a new classification task with a set of classes $C_{eval}$ where $|C_{eval} - C_{train}| \geq 1$. In such cases, the last layer needs to be replaced with number of units equal to $C_{eval}$ and the model parameters needs to be updated using set of some samples $\mathcal{S} = \{(\mathbf{x}_1, y_1), (\mathbf{x}_2, y_2), \dots (\mathbf{x}_K, y_K)\}$ from the new task and then the performance is evaluated on the set of unseen samples of that task $\mathcal{Q} = \{(\mathbf{x}^*_1, y^*_1), (\mathbf{x}^*_2, y^*_2), \dots (\mathbf{x}^*_T, y^*_T)\}$. Here $\mathcal{S}$ and $\mathcal{Q}$ are known as the \textit{Support Set} and   \textit{Query Set} respectively. When, $|C_{eval}| = N$ and the number of samples per class, $k_c = k$, we refer the task as "$N$-way, $k$-shot" classification task. When a model is evaluated on numerous such tasks independently, we call it episodic evaluation, where each task is referred to as an episode.

\subsection{Episodic Learning Approaches}
\label{subsec:episodic_learning_approaches}

There exist many episodic learning approaches, out of which we focus mainly on meta-learning algorithms. Meta-learning is the ability of a model (or learner) to gain experience across multiple tasks (or episodes), such that, given a new episode with a very few samples, it quickly adapts. Here, we select episodic gradient-based meta-learning approaches, specifically Model-Agnostic Meta-Learning (MAML)~\cite{finn2017model} and its variants. In the context of gradient-based methods, the term \textit{adapting quickly} refers to a few parameter updates. Here, we briefly describe the meta-learning approaches used in this work:

\textbf{MAML~\cite{finn2017model}}: Given an episode with $K$ classes and an embedding function $f(.,\theta)$, a learnable linear layer $\textit{clf}(.,\phi)$ is added on it and trained on the support set of the episode via gradient descent iterations, and evaluated on the query set. This step is performed independently for each episode, and the loss on the query set across all training episodes is aggregated. Which is then backpropagated through within-episode iterations to compute gradients of the query loss with respect to $\theta$ and $\phi$. These gradient is then used to update the parameters, and this is referred to as the meta-update. However, this involves computing second-order derivatives, which can be computationally expensive. Hence, we use its first-order approximation, also known as fo-MAML~\cite{finn2017model}, where the within-episode gradients are ignored during the meta-update.

\textbf{Proto-MAML~\cite{triantafillou2019meta}}: Proto-MAML is inspired by Prototypical networks~\cite{snell2017prototypical}, where a prototype $\mathbf{p_i}$ for class $c_i$ is the average of the embeddings of the samples of the class $c_i$ in the support set. A query sample is classified as class $c_i$ if the embedding of the sample is nearest to its prototype $\mathbf{p_i}$ in the Euclidean space. \citet{snell2017prototypical} explains that prototypical networks can be re-interpreted as a linear layer applied on the embeddings from $f(.,\theta)$ where the weights for the unit representing the class $c_i$ is $\mathbf{w_i} = 2\mathbf{p_i}$ and bias is $b_i =-||\mathbf{p_i}||^2$. Proto-MAML uses this assignment of $\mathbf{w_i}$ and $b_i$ to initialize the parameters of $\textit{clf}(.,\phi)$. Here, we use the fo-MAML with the prototypical assignment of parameters of $\textit{clf}(.,\phi)$ and refer to this version as proto-fo-MAML.
\section{Related Work}
\label{sec:related_work}

Detecting observed or imagined text from EEG has been actively studied in the literature. Some earlier works have achieved good performance in EEG-to-Text decoding with closed-vocabulary datasets containing very few classes~\cite{saha2019speak,nieto2022thinking}. Recent works focus on open vocabulary EEG-to-Text decoding with thousands of classes, where they either attempt to reconstruct~\cite{wang2022open,liu2024eeg2text,tao2025see} or semantically summarize~\cite{liu2025learning} the text a person views on a screen from the EEG signals. However, an EEG might not capture representations of each word, including all parts of speech, but it might capture representations of nouns corresponding to real objects. It is known that the human brain processes visual objects hierarchically~\cite{cichy2014resolving,carlson2013representational}. Here, we explore whether representations of objects in the brain induced via textual stimuli of the object follow a similar hierarchy.

As far as hierarchical modelling is concerned, the works of \citet{zhu2024eeg} and \citet{triantafillou2019meta} are closely related to our work. \citet{zhu2024eeg} created EEG-ImageNet, which is a dataset of EEG and image pairs across $80$ classes, where $40$ classes are coarse-grained, distinct classes from ImageNet~\cite{deng2009imagenet}, whereas the other $40$ classes come from $5$ groups corresponding to concepts in WordNet with $8$ categories each. However, EEG-ImageNet exhibits a very limited hierarchy comprising only $5$ concepts from WordNet. In contrast, our test split contains $75$ concepts, enabling a rigorous hierarchical study. \citet{triantafillou2019meta} performs a similar strategy as ours to build a hierarchy. However, our goal differs fundamentally in terms of using hierarchy-aware episodes as a probing tool to analyze neural representations under varying ontological abstraction, rather than training general-purpose few-shot learners. We also perform rigorous hierarchical studies given a rich hierarchy in the test split, unlike~\cite{triantafillou2019meta}.

Regarding episodic Learning in the EEG decoding literature, most studies focus on cross-subject few-shot learning. As the EEG signal across subjects varies widely, they address this issue by treating each subject as an episode and applying episodic meta-learning, enabling the models to adapt quickly to novel subjects~\cite{han2024meta,chen2025model,duan2020meta,bhosale2022calibration}.
In most cases, episodes have a limited labels that remain fixed across episodes. In a realistic scenario, this might not be the case; hence, we need a more realistic organization of episodes, for which we adopt a hierarchical structure.

\section{Hierarchic-EEG2Text}
In this section, we elaborate on how we build the framework for assessing EEG-to-Text decoding across hierarchical abstraction levels. We first briefly describe the dataset and the preprocessing pipeline, then describe building the concept hierarchy, and finally describe the episode sampling strategy.
\label{sec:hierarchical_EEG2Text}
\subsection{Dataset}

We used the PEERS dataset~\cite{ds004395:2.0.0} for our experiments. The PEERS dataset consists of EEG signals recorded while subjects observed English words displayed on the screen (and later recalled them) under varying conditions in which they performed decision-making and arithmetic tasks, thereby inducing signal variability. The dataset consists of five major experiments grouped into $3$ sections: \textit{ltpFR}, \textit{ltpFR2}, and \textit{VFFR}, in which $300+$ subjects contributed $7000+$ sessions of $90$ minutes each, observing words from a pool of $1638$ English words. In this work, we have used \textit{ltpFR} and \textit{ltpFR2} groups, whose descriptions are provided in appendix \Cref{sec:dataset}.

\subsubsection{Alignment of Multiple Montage}
\label{subsubsec:montage_alignment}

EEG signals in the PEERS dataset were recorded using $3$ different electrode layouts: (i) 129-channel GSN 200, (ii) 129-channel HydroCel GSN, and (iii) 128-channel BioSemi headcap. In this work, our main objective is to perform a hierarchical study of EEG-to-Text, rather than montage-agnostic EEG processing. Although we use a recent montage agnostic architecture, for a fair comparison with traditional EEG architectures, we select $96$ channels so that all EEG data can be used in the study. For channel selection, we use one of the layouts as the reference layout and apply $k$-nearest neighbor to find the corresponding channel in the other layout. We do so because not all subjects have EEG electrode placement in exactly the same way. There will always be some variability in electrode placement across subjects and sessions due to differences in head shape and manual headset use. An electrode $x$ may sometimes be placed closer to its neighbor $y$ and sometimes farther apart. The detailed steps involved in the selection of the channels are outlined as follows:

\textbf{Step 1:} We consider the BioSemi ABC layout (used by the 128-channel BioSemi headcap) as the reference layout. For each electrode in the reference, we find $8$-nearest neighbors in the other two (target) layouts using the known $xyz$-location of electrodes. This way we get a tuples in the form of $(e^{r}_i, \mathbf{e^t_i})$ where $e^{r}_i$ is an electrode in the reference layout and $\mathbf{e^t_i}$ is a list of neighbors in the target layout such that $\mathbf{e^t_i}_j$ is the $j^{th}$ nearest neighbor of $e^{r}_i$.   

\textbf{Step 2:} There might be a case where two electrodes in the reference layout have the same nearest neighbor. In other words, for $i\neq j$, there may exist two tuples $(e^{r}_i, \mathbf{e^t_i})$ and $(e^{r}_j, \mathbf{e^t_j})$ such that $\mathbf{e^t_i}_1 = \mathbf{e^t_j}_1$. We use the Euclidean distance $d(.,.)$ to resolve such conflicts, and repeat this step $8$ times. If the nearest-neighbor list is exhausted to only $1$ electrode, we skip that tuple during conflict resolution. If there are still duplicates, only one is retained.

\textbf{Step 3:} After de-duplication at step $2$, we select all tuples such that there exists a nearest neighbor for a reference electrode in both the layouts. 

\subsubsection{Pre-processing}

As we also study architectures pre-trained on the TUEG corpus~\cite{obeid2016temple,wang2024cbramod}, we follow pre-processing steps similar to those used in~\cite{wang2024cbramod}, \textit{i.e.}, (i) re-reference the data to the average, (ii) extract electrodes as per selection in \Cref{subsubsec:montage_alignment} based on the layout, (iii) downsample to $200$Hz, (iv) apply a band-pass filter from $0.3-75$Hz, (v) set the unit to $100 \mu V$. We used $1$-sec EEG signal window for each word in the dataset.

\subsection{Hierarchy of Concepts}
\label{subsec:hierarchy_of_concepts}

In order to create a hierarchy of concepts, we used NLTK's WordNet~\cite{bird2006nltk,princetonuniversity_2019_wordnet} interface to build a Directed Acyclic Graph (DAG) using the \textit{IS-A} relationship. Key steps involved in the creation of DAG are as follows:

\textbf{Intializing the DAG.} We consider all the words in the PEERS Dataset~\cite{ds004395:2.0.0} as the leaves of the DAG. For each leaf node, we use the closest synonym from the synset returned by WordNet as the parent of the leaf node. Then, we use the hypernym-path from the root in WordNet to the given synonym to create the path from the root to the leaf. We discard all words whose synsets are not available in WordNet, resulting in a DAG with a single root, where each leaf node corresponds to a word in the dataset and each internal node represents a WordNet concept.

\textbf{Removal of Broad Words.} The set of words in the PEERS dataset contains several words that are too broad (\eg \textit{ animal, mammal, creature, and so on}) and are placed close to the root. In our experiments, we considered words with narrower meanings and discarded broader ones similar to WordNet concepts. To identify such words, we compute the total number of siblings of each leaf node and subtract this number from the total number of leaf nodes reachable by the parent of the corresponding leaf node. If the result is too high ($\geq 45$), the word is possibly too broad. We remove such words from the DAG (check appendix \Cref{sec:app_dicarded_labels} for related details).

\textbf{DAGs for Train-Val-Test Split.} For \textit{meta-train} we use data from ``ltp\_FR'' and for \textit{meta-validation} and \textit{meta-test}, we use the 576 words from the ``ltp\_FR2'' group of the PEERS dataset. This choice ensures that each task has sufficient samples in both the support and query sets during episodic evaluation, as words in the ``ltp\_FR2'' group are shown to each subject more times than those in the ``ltp\_FR'' group. We retain all \textit{(word, subject)} pairs with $\geq 23$ occurrences and remove these pairs from the \textit{meta-train} split. The selected words and subjects are then randomly divided between the \textit{meta-validation} and \textit{meta-test} splits. Although initializing the DAG with all words helped filter out overly broad concepts, we reconstruct the DAG separately for each of the \textit{meta-train}, \textit{meta-validation}, and \textit{meta-test} based only on the words assigned to that split. This results in a \textit{meta-train} split with $1126$ words, $204$ subjects, and $484832$ samples; a \textit{meta-validation} split with $92$ words, $15$ subjects, and $32393$ samples; and a \textit{meta-test} split with $392$ words, $45$ subjects, and $414313$ samples.

\subsection{Episode Sampling}
\label{subsec:episode_sampling}

We use the episode sampling method similar to that in ~\cite{triantafillou2019meta}. We present the key steps as follows:

\textbf{Sampling Classes.} For each DAG, we first identify the set of eligible internal nodes. We discard nodes that are too close to the root (see appendix \Cref{sec:app_dicarded_labels}) and retain internal nodes having at least $5$ reachable leaf nodes. We then uniformly sample one internal node from this eligible set and treat the leaf nodes reachable from it as the candidate classes. If the number of reachable leaf nodes is $\geq 10$, we randomly select $10$ words as the classes; otherwise, we use the entire set of reachable leaves. We refer to the set of reachable leaves from an internal node as the "span" of the node, and to the span's size as the "span length".

\textbf{Shots for Query Set.} Number of samples per class for the query set is kept constant across all episodes and is determined by \Cref{eq:q_per_class}.

\begin{equation}
    \label{eq:q_per_class}
    k_{q} = \min\Bigl\{10, \Bigl(\min_{c \in \mathcal{C}} \lfloor 0.5*|\mathcal{S}(c)| \rfloor\Bigr)\Bigr\}
\end{equation}

where $\mathcal{S}(c)$ is the set of samples available for class $c$ and $\mathcal{C}$ is set of classes sampled for the episode. \Cref{eq:q_per_class} ensures that at least half the samples for all classes considered in the episode are available for the support set.

\textbf{Total Support Size.} To avoid a too large support set, the number of samples for each class is limited to at most 100. Specifically, the total support size is determined by \Cref{eq:support_size}. 

\begin{equation}
    \label{eq:support_size}
    |\mathcal{D}_{sup}| = \min \Bigl\{100, \sum_{c \in \mathcal{C}} \lceil \beta*\min\{100, |\mathcal{S}(c)| - k_{q}\} \rceil \Bigr\}
\end{equation}

where $\beta$ is uniformly sampled from $(0,1]$, which ensures episodes of both large and small support sets.

\textbf{Shots for Support Set.} First, the proportion of available samples per class that would contribute to the support set is determined by the following normalized exponential function:

\begin{equation}
    R_c = \frac{\exp(\alpha_c + \log(|\mathcal{S}(c)|))}{\sum_{c' \in \mathcal{C}} \exp(\alpha_{c'} + \log(|\mathcal{S}(c')|))}
\end{equation}

where $\alpha_c$ is uniformly sampled from $[\log(0.5), \log(0.2))$ which brings in some noise in the distribution. Finally, the shots per class are determined as:
\begin{equation}
    k_{{sup}^c} = \min \{\lfloor R_c*(|\mathcal{D}_{sup}|-|\mathcal{C}|) \rfloor + 1, |\mathcal{S}(c)| - k_q\}
\end{equation}

While sampling episodes for episodic training, we ensure that the support and query sets are disjoint not only within each episode but also across all episodes.

During episodic evaluation, we iterated through all the eligible internal nodes, and for each such node, we sampled $\mathcal{I}$ instances of episodes to ensure most leaves from the span are considered, where $\mathcal{I}_{meta-validation} = 5$ and $\mathcal{I}_{meta-test} = 10$. We report the mean and standard deviation of performances across all internal nodes.

\subsection{Metrics}

For all experiments, we have reported the "chance-adjusted accuracy" or the "normalized accuracy" for classification tasks, so that the evaluation scale remains the same across $N$-way classification tasks with varying $N$. The normalized accuracy for an $N-way$ classification task is defined as:

$$
\textit{Normalized Accuracy} = \frac{A-\frac{1}{N}}{1-\frac{1}{N}} * 100
$$

Where $A$ is either accuracy or balanced accuracy. A normalized accuracy of $0$ is as poor as a random guess, while a negative value indicates performance poorer than a random guess. The closer the value is to $100$, the better.
\section{Experiments and Results}
\label{sec:experiments}

\begin{figure*}[!t]
  \begin{center}
    \centerline{\includegraphics[width=\linewidth]{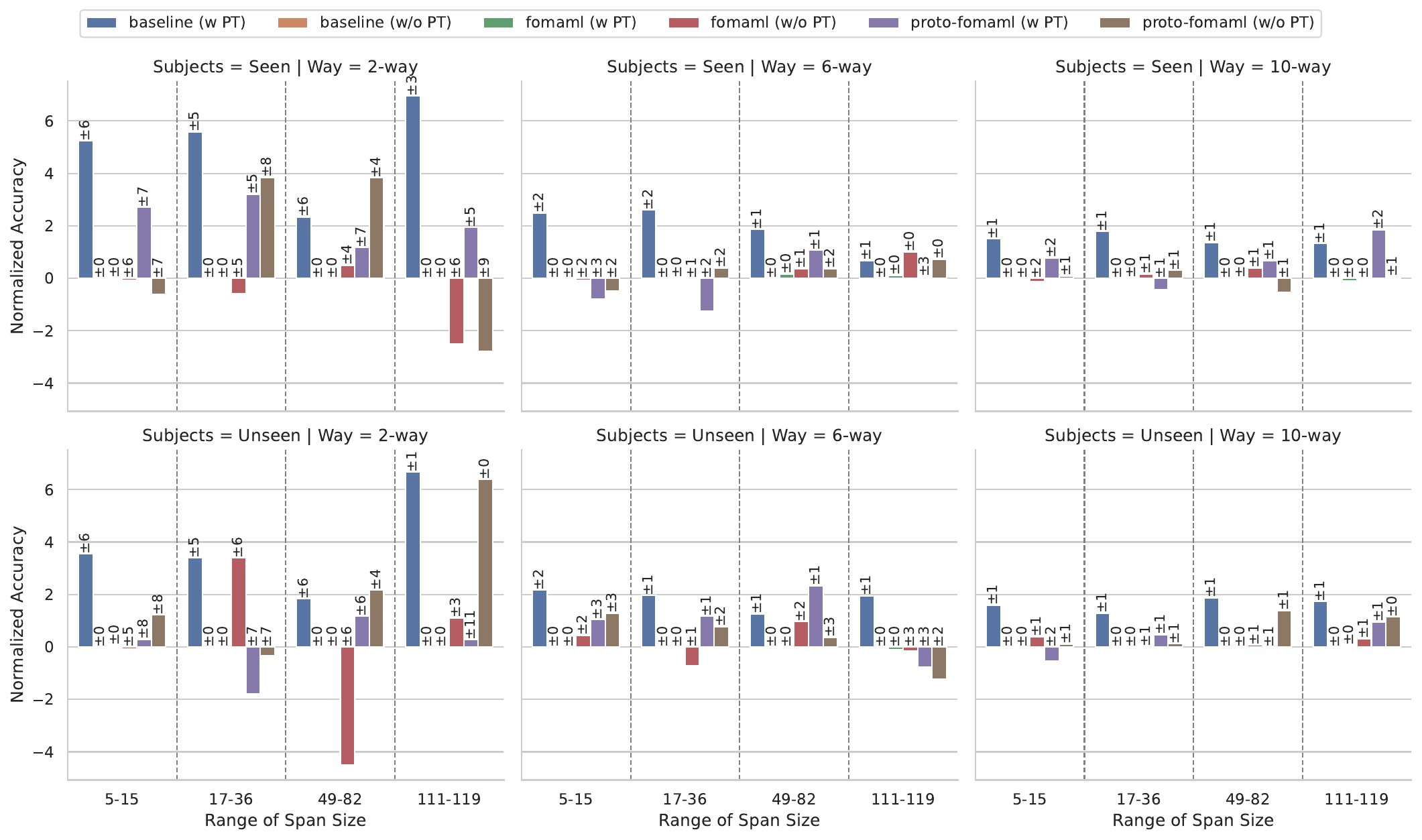}}\vspace{-1.0em}
    \caption{
      Span Analysis for episodic evaluation of CBraMod. It depicts the effects of an episode's hierarchical depth on performance. Higher span size indicates coarse-grained episodes, and lower span size indicates fine-grained episodes.
    }
    \label{fig:span_analysis_combined}
  \end{center}\vspace{-1.00em}
\end{figure*}

Our primary objective is to study EEG-to-Text decoding at various hierarchical levels. Through our experiments, we show that performing non-episodic evaluation with a large number of classes can be difficult for DNN models and that hierarchical episodic evaluation yields insights into various levels of abstraction. We perform our experiments on three architectures: (i) EEGNet~\cite{lawhern2018eegnet} is a traditional EEG decoding architecture mainly built of convolution layers, (ii) NICE-EEG~\cite{song2023decoding} is a modern EEG decoding architecture built with a spatial transformer followed by convolution layers, and (iii) CBraMod~\cite{wang2024cbramod} is a foundational model pretrained on a large EEG corpus. We used CBraMod, both with pre-trained (w PT) and without pre-trained (w/o PT) weights, to study the effect of self-supervised pretraining. 

\subsection{Episodic Evaluation}
\label{subsec:episodic_eval}

For episodic evaluation, we trained the architectures in two different setups covering three algorithms: (i) Baseline, where we train the architectures with all the classes of \textit{meta-train} together as a typical classification task in a non-episodic manner, but perform validation and evaluation in an episodic manner as described in \Cref{subsec:episode_sampling}, (ii) Episodic learning, where we train the architectures using fo-MAML and proto-fo-MAML.

\Cref{tab:few_shot_unseen} specifies overall episodic performance for $2-way$, $6-way$, and $10-way$ episodes. This also reflects the episodic performance on unseen classes and subjects, as the sets of classes and subjects for \textit{meta-train}, \textit{meta-val}, and \textit{meta-test} are disjoint. The key observations are that object categorization from brain-wide EEG activity during real-world tasks is very challenging, and that there is modest improvement in performance compared to non-episodic evaluation, as shown in~\Cref{tab:non_episodic_eval}, suggesting better performance in some episodes. The near-chance performance, specifically on $6-way$ and $10-way$ classification tasks, could be due to the greater complexity and variability of EEG signals when subjects engage in real-world reasoning. In some trials, the words were displayed without additional tasks, while in others, decision tasks involving animacy and size properties were included. This highlights the need for further advances in EEG decoding architectures to tackle such variability. 
\begin{table*}[!t]
\centering
\caption{Normalized Accuracy for episodic evaluation on unseen subjects}\vspace{-1.0em}
\label{tab:few_shot_unseen}
\begin{sc}
\resizebox{0.8\textwidth}{!}{%
\begin{tabular}{@{}l|lll|lll|lll@{}}
\toprule
Embedder           & \multicolumn{3}{l|}{Baseline} & \multicolumn{3}{l|}{fo-MAML~\cite{finn2017model}} & \multicolumn{3}{l}{proto-fo-MAML~\cite{triantafillou2019meta}} \\ \cline{2-10}
                   & 2-way   & 6-way   & 10-way   & 2-way   & 6-way   & 10-way  & 2-way     & 6-way     & 10-way    \\ \midrule
EEGNet~\cite{lawhern2018eegnet}           & 0.91\textpm 2.6 & 0.40\textpm 1.1 & 0.57\textpm 0.8  & 0.02\textpm 0.2       & -0.02\textpm 0.1   & 0.00\textpm 0.1     & 0.14\textpm 0.5     & 0.04\textpm 0.4 & 0.00\textpm 0.3   \\
NICE EEG~\cite{song2023decoding}         & 2.08\textpm 4.0 & 1.18\textpm 1.1 & 0.88\textpm 1.1  & 0.02\textpm 0.7 & 0.10\textpm 0.4   & 0.00\textpm 0.4     & 0.12\textpm~ 0.8 & 0.03\textpm 0.4 & -0.06\textpm 0.5  \\
CBraMod~\cite{wang2024cbramod} (w/o pt) & 0.00\textpm 0.0       & 0.00\textpm 0.0       & 0.00 \textpm 0.0       & 0.28\textpm 5.3 & 0.22\textpm 2.1  & 0.22\textpm 1.1  & 1.22\textpm 7.1  & 0.99\textpm 2.8 & 0.36\textpm 1.30    \\
CBraMod~\cite{wang2024cbramod} (w pt)   & 3.53\textpm 6.0 & 2.04\textpm 1.4 & 1.54 \textpm 1.0 & 0.04\textpm 0.4 & -0.04\textpm 0.2 & -0.03\textpm 0.1 & -0.01\textpm 7.6 & 1.09\textpm 2.3 & -0.02\textpm 1.4  \\ \bottomrule
\end{tabular}
}
\end{sc}\vspace{-0.5em}
\end{table*}
\begin{table*}[!t]
\centering
\caption{Normalized Accuracy for episodic evaluation on seen subjects}\vspace{-1.0em}
\label{tab:few_shot_seen}
\begin{sc}
\resizebox{0.8\textwidth}{!}{%
\begin{tabular}{@{}l|lll|lll|lll@{}}
\toprule
Embedder           & \multicolumn{3}{l|}{Baseline} & \multicolumn{3}{l|}{fo-MAML~\cite{finn2017model}} & \multicolumn{3}{l}{proto-fo-MAML~\cite{triantafillou2019meta}} \\ \cline{2-10}
                   & 2-way   & 6-way   & 10-way   & 2-way   & 6-way   & 10-way  & 2-way     & 6-way     & 10-way    \\ \midrule
EEGNet~\cite{lawhern2018eegnet}           & 1.27\textpm 2.3 & 0.5\textpm 0.9  & 0.49\textpm 0.6 & -0.03\textpm 0.3  & -0.02\textpm 0.1 & -0.01\textpm 0.1  & 0.09\textpm 0.7  & 0.01\textpm 0.4 & 0.06\textpm 0.3 \\
NICE EEG~\cite{song2023decoding}         & 2.36\textpm 3.0 & 1.02\textpm 1.4 & 0.84\textpm 0.9 & -0.11\textpm 0.7 & 0.01\textpm 0.4  & -0.06\textpm 0.2 & -0.34\textpm 1.1 & -0.11\textpm 0.6 & 0.09\textpm 0.5 \\
CBraMod~\cite{wang2024cbramod} (w/o pt) & 0.00\textpm 0.0 & 0.00\textpm 0.0 & 0.00\textpm 0.0 & -0.22\textpm 5.8 & 0.03\textpm 1.6  & 0.04\textpm 1.3  & 0.37\textpm 6.9  & -0.18\textpm 2.4 & 0.08\textpm 1.2 \\
CBraMod~\cite{wang2024cbramod} (w pt)   & 5.17\textpm 5.5 & 2.38\textpm 1.9 & 1.57\textpm 0.9 & 0.01\textpm 0.1  & 0.00\textpm 0.2  & -0.02\textpm 0.2 & 2.66\textpm 6.6  & -0.70\textpm 2.5 & 0.44\textpm 1.6 \\ \bottomrule
\end{tabular}
}
\end{sc}
\end{table*}
Secondly, the baseline setup outperforms both meta-learning algorithms, despite their designs for multi-task learning that involve such variability. Another key observation is that CBraMod with pretrained weights performs best, while CBraMod trained from scratch without pretrained weights performs worst due to overfitting, indicating the benefit of large-scale self-supervised pretraining. The consistent advantage of pretrained baselines over meta-learning suggests that representation quality dominates adaptation in high-noise EEG settings. Meta-learning may amplify noise when task boundaries are weakly defined. Similar results have been found for episodic learning for images, where a self-supervised pretrained model performs better than sophisticated meta-learning algorithms~\cite{tian2020rethinking}.

Beyond these reasons, we want to emphasize again that the stimuli and tasks in the PEERS dataset reflect real-world reasoning and are also closer to clinical applications. Brain activity in PEERS is likely to be more abstract and complex than that of image/video EEG datasets (\eg\ SEED~\cite{seedazheng2015investigating,seedbduan2013differential}, SEED-IV~\cite{seediv8283814}, EEGCVPR40~\cite{spampinato2017deep}, ThoughtViz~\cite{kumar2018envisioned,kaneshiro2015representational}), etc, where visual stimuli evoke wider and more stereotyped brain activity~\cite{yamins2014performance} and benefit ML models for categorization. We believe that our evaluation brings forth the challenges of meta-learning on EEG datasets with more complex tasks.

We also evaluated performance on $9$ seen subjects from the \textit{ltpFR} group of the PEERS dataset, which had classes included in the \textit{meta-test}. Although these subjects were observed by the model during training, the set of classes still remained disjoint. \Cref{tab:few_shot_seen} specifies the episodic performance for the seen subjects. While most observations remain the same as for unseen subjects, overall performance increased specifically for CBraMod with pretrained weights.

Another key observation here is that the standard deviation is quite high in the performance aggregated across all the internal nodes, which indicates that the model may have performed better for some episodes/concepts, while for some, it performed poorer, which highlights the need for more fine-grained analysis, which we discuss in the next section. 

\subsection{Hierarchical Analysis}

The results in the previous section provide a coarse-grained view of the EEG-to-Text decoding across episodes. Here, we attempt to provide a fine-grained view via: (i) Span Analysis, where we study performance at different depths of concepts, and (ii) Abstraction Analysis, where we pick classes from different levels of the hierarchy DAG. For the upcoming sections, we use CBraMod due to its superior performance.

\subsubsection{Span Analysis}

\begin{figure*}[]
  \begin{center}
    \centerline{\includegraphics[width=\linewidth]{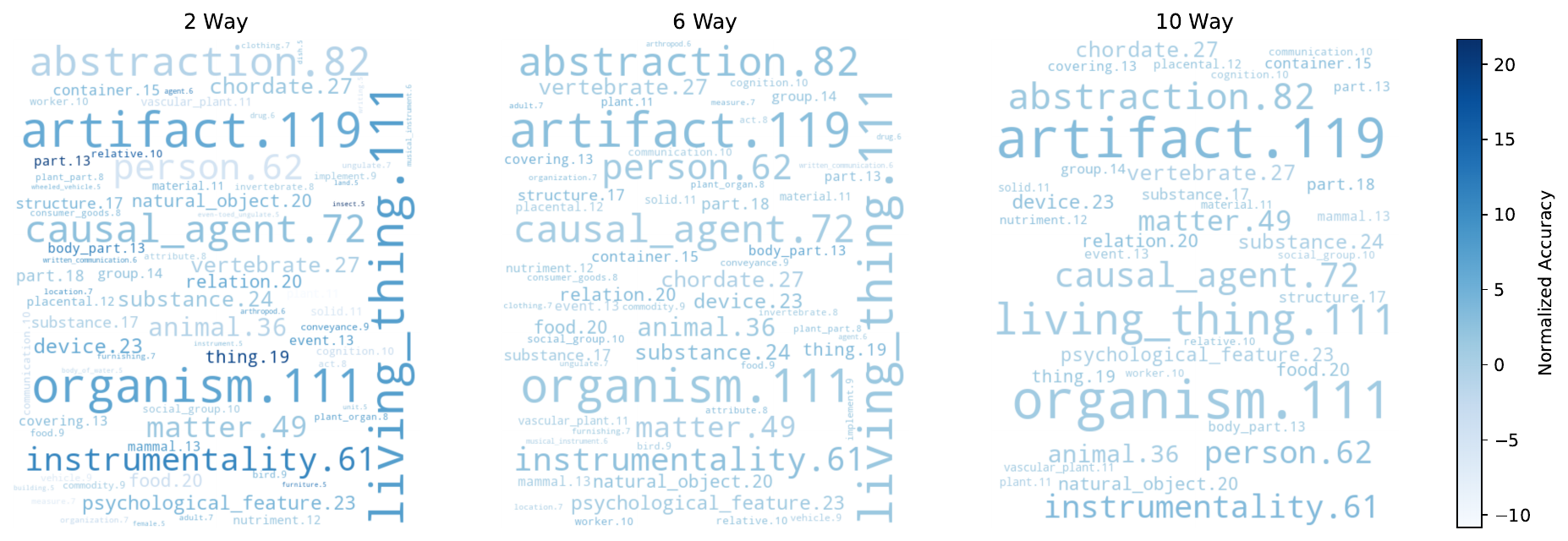}}\vspace{-1.0em}
    \caption{
      Fine-grained Span Analysis for episodic evaluation of CBraMod on unseen subjects (w PT) trained via baseline learning algorithm. The word cloud represents the concepts represented by the internal nodes in the DAG of the \textit{meta-test} split. Word size and numeric suffix represent the span size of the node, and brightness represents the normalized classification accuracy.
    }
    \label{fig:word_cloud_combined}
  \end{center}\vspace{-1.5em}
\end{figure*}

As defined earlier, we refer to the set of reachable leaves from an internal node as the "span" of that node in the hierarchy DAG, and to the span's cardinality as the "span length". The span of an internal node also specifies how close the node is to the root (how broad is the meaning of the concept) and how far the node is from the root (how narrow is the meaning of the concept). The broader the span size, the broader the concept is. We use span instead of the number of hops, as the DAG might not be balanced. \Cref{fig:span_analysis_combined} visualizes the performance as we vary the span size of an internal node in the DAG of \textit{meta-test} for unseen and seen subjects. As the span size ranges from $5$ to $119$ across the DAG, we divide them into four divisions: (i) Group of very narrow nodes having span size from $5$ to $15$, (ii) Group of moderately narrow nodes having span size from $17$ to $36$, (iii) Group of moderately broad nodes having span size from $49$ to $82$, and (iv) Group of broad nodes having span size from $111$ to $119$. We can observe from \Cref{fig:span_analysis_combined} that baseline learning shows consistent behavior when pretrained weights are used, whereas other learning methods remain unstable, often performing below chance. One key observation from the consistent behavior of the baseline learning is that the performance for the broader concepts (having span size ranging from $111$ to $119$) remains the highest for $2-way$ classification. This could be because sampling two classes from a broad span set could return very distinct classes. Coarse-grained classification seems easier than fine-grained classification. But this might be true only for $2-way$ classification and not in general, as observed by the dip in performance for $6-way$ and $10-way$, which may be because of the presence of more classes of very similar meaning, making classification difficult (check \Cref{fig:sub_graph} in the appendix for an example). Similar behavior can be observed for the narrower (having span size from $5$ to $15$) and moderately narrower nodes (having span size from $17$ to $36$), where the performance remains better for $2-way$ classification, but when we put more classes as in $6-way$ and $10-way$ classification tasks, the performance dips.

\Cref{fig:word_cloud_combined} provides a more fine-grained analysis of the variation of accuracy vs the span size via word clouds. The numeric suffix and size of words (WordNet concepts) in the word cloud represent the span size \ie, the larger the word, the broader the node or concept. The brightness of the words represents the performance. The brighter the words, the higher the performance. The word cloud visualization gives a clear view of the observation in \Cref{fig:span_analysis_combined} for the unseen subjects. Broader concepts (like \verb|artifact.119|, \verb|organism.111|, \verb|living_thing.111|) perform well in $2-way$ classification, but their performance dips for $6-way$ and $10-way$ as the brightness of the concepts in the word cloud dips. Another key observation from \Cref{fig:span_analysis_combined} is that the performance for CBraMod (w PT) for baseline learning is the lowest for the moderately broader concepts in $2-way$ task, suggesting the presence of very similar classes in the span set. As per \Cref{fig:word_cloud_combined}, this group includes concepts like \verb|causal_agent.72|, \verb|person.62|, \verb|matter.49| that have very similar classes. For example, the intersection of the span set of \verb|causal_agent.72| and \verb|person.62| contains classes like ACTOR, ACTRESS, HOSTESS, GIRL, DAUGHTER, DANCER, etc, which just depict an image of a person. Similarly, most words in the span set of \verb|matter.49| represents some kind of food items (check appendix \Cref{sec:span_set_lists} for full list), and differentiating such items via EEG representations can be a difficult task, which explains the low performance.

\begin{table}[!t]
\centering
\caption{Normalized (balanced) accuracy on joint evaluation with 1126 classes, together with a stratified train-val-test split.}\vspace{-1.0em}
\label{tab:non_episodic_eval}
\begin{sc}
\begin{tabular}{@{}lll@{}}
\toprule
Embedder         & Train         & Test         \\ \midrule
EEGNet~\cite{lawhern2018eegnet}           & 0.03    & 0.00  \\
NICE EEG~\cite{song2023decoding}         & 53.24  & 0.01  \\
CBraMod~\cite{wang2024cbramod} (w/o pt) & 0.00    & 0.00   \\
CBraMod~\cite{wang2024cbramod} (w pt)   & 1.55   & 0.02   \\ \bottomrule
\end{tabular}%
\end{sc}
\end{table}

\subsubsection{Abstraction Analysis}

\begin{figure*}[!t]
  \begin{center}
    \centerline{\includegraphics[width=1.0\linewidth]{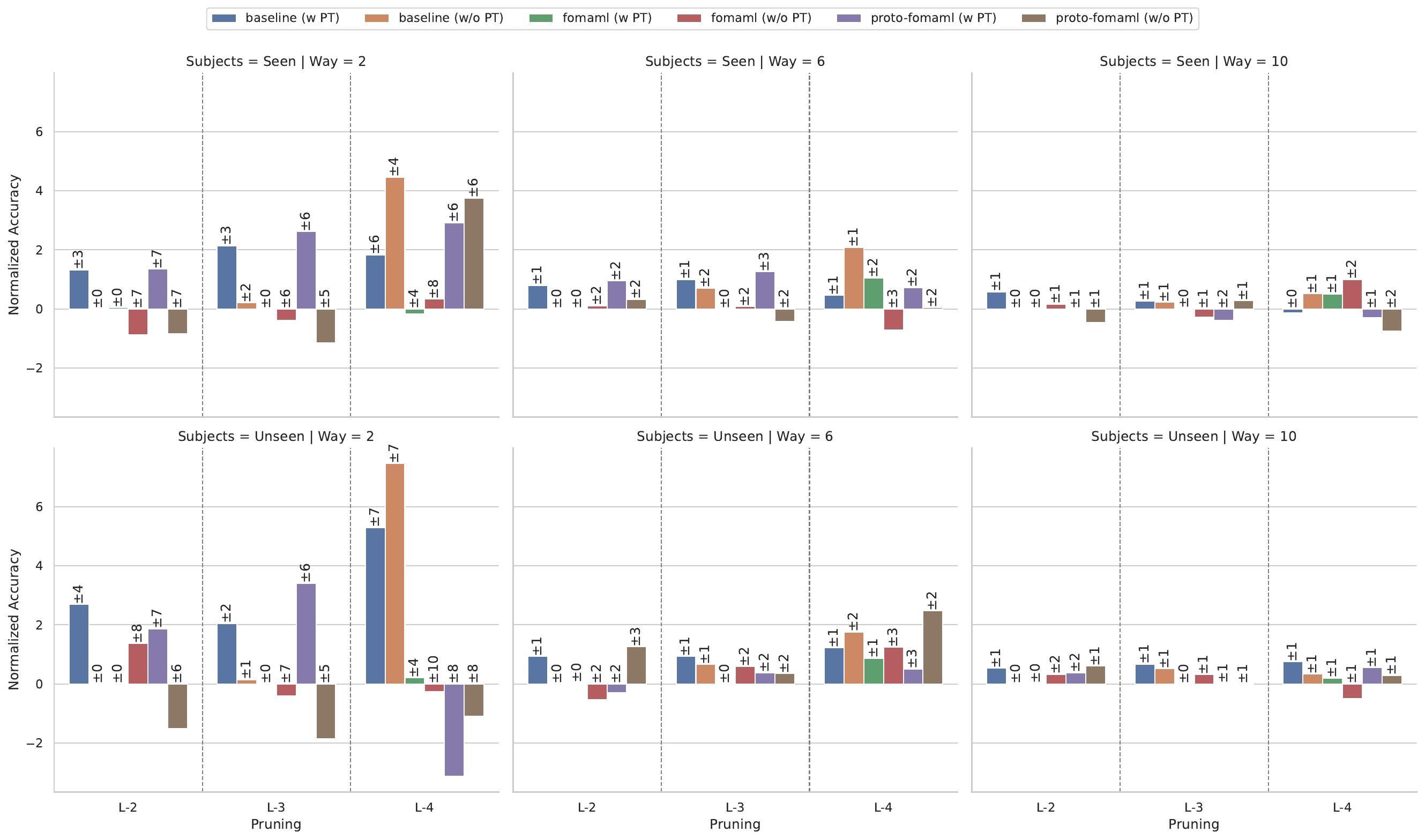}}\vspace{-1.0em}
    \caption{
      Abstraction Analysis for episodic evaluation of CBraMod. Value of $h$ in $L-h$ specifies the number of pruning steps performed on the DAG. As the value of $h$ increases, more abstract classes are considered.
    }
    \label{fig:abstraction_analysis_combined}
  \end{center}\vspace{-1.5em}
\end{figure*}

In this section, we study how classes at different levels of the hierarchy contribute to EEG-to-Text decoding. In all the previous sections, we used the leaf node of the DAG as the classes in the classification task. Here, we select the concepts represented by internal nodes that are ancestors of the leaf nodes with $h$ hops in between them. If $h=1$, then we prune all the leaf nodes and consider the new leaves as the classes, and we refer to the new set of classes as $L-1$. If $h=2$, we prune twice and refer to the resulting set of classes as $L-2$. After pruning, we assign samples to the new classes using the ancestor-descendant mapping between the old and new leaves. Towards this end, we considered three sets of classes: $L-2$, $L-3$, and $L-4$. For each set, we trained the model independently using all learning methods and evaluated performance episodically. An important point to note is that when classes are selected from the leaves without any pruning, the classes are disjoint across \textit{meta-train}, \textit{meta-validation}, and \textit{meta-test}. However, when we prune the DAG, the set of classes begins to overlap across the splits. Specifically, $\sim25\%$ of test classes overlapped with that of the train classes for all three sets $L-2$, $L-3$, and $L-4$. \Cref{fig:abstraction_analysis_combined} visualizes the episodic performance of CBraMod on the three sets trained with various learning approaches. As in earlier results, the baseline learning method performs consistently, while meta-learning remains unstable and shows no trend. However, for the baseline method, overall performance increases in most cases. Specifically, when CBraMod is trained without pretraining, its performance improves at $L-4$. Moreover, it beats the case when CBraMod is initialized with pretrained weights. This observation yields two insights: (i) the possibility of more distinct classes as we move a few hops above the last level of the hierarchy, and (ii) improvement in the results at $L-4$ indicates that EEG representations are sensitive to levels of abstraction in the ontological hierarchy.

We emphasize that absolute performance remains near chance, and variance is high across episodes. Our goal is not to demonstrate strong decoding accuracy, but to analyze relative trends across controlled hierarchical conditions under identical evaluation protocols.

\subsection{Non-Episodic Evaluation}


In this experiment, we train and evaluate the architectures in a non-episodic manner, where we use the data under \textit{meta-train} and divide them into stratified \textit{train-validation-test} splits (note that \textit{meta-train}, \textit{meta-val}, and \textit{meta-test} splits are meant for episodic learning only). We train the architectures to map EEG samples to $1126$ classes, and the set of classes remains the same for both training and evaluation, as described in \Cref{subsubsec:non-episodic-eval}. The \Cref{tab:non_episodic_eval} specifies the performance of the architectures in the train and test sets in terms of normalized balanced accuracy (mean of recall for all classes). 

We observe that all architectures fail to generalize to the test set. This highlights the difficulty of classifying signals into a large set of classes. One possible reason is that certain words (or leaf nodes) under the same internal node in the DAG may share high semantic similarity (see \Cref{fig:sub_graph} in the appendix for an example). EEG might not capture representations for each of them, and multiple such instances make the classification task difficult when all classes are considered together. On the other hand, as we show in \Cref{subsec:episodic_eval}, performing an independent classification task for such nodes by treating their span sets as classes may yield different results, as the span sets of some nodes may contain more distinct objects compared to others.
\section{Limitations and Ethical Considerations}

In this work, we focus on hierarchical analysis of EEG-to-Text decoding to understand how EEG captures object representations in the brain that correspond to textual stimuli. We used the PEERS~\cite{ds004395:2.0.0} dataset, which contains large classes of real objects. However, even the most modern architecture, such as CBraMod, performed close to chance level, according to our findings. 

This may have occurred because of high variability in the dataset, as the subjects also performed additional tasks while viewing words corresponding to objects. This implies that the current methods fail to capture representations from EEG alone and may require support from other modalities. In this direction, it would be interesting to study how aligning EEG representations with other modalities, such as textual representations, would help. As a part of future work, we plan to explore this direction to provide more robust insights. Additionally, we will explore other vision-based datasets on which ML models have achieved higher accuracy, to evaluate the benefits of meta-learning and ontological priors.

Regarding ethical considerations, we haven't recorded physiological signals; instead, we used a publicly available dataset. Information about subjects is already anonymized by the dataset authors, and we ensure that our methodology doesn't extract any information that can be used to identify subjects.
\section{Conclusion}
\label{sec:conclusion}

In this work, we study EEG-to-Text decoding across hierarchical abstraction levels. We show that classifying EEG signals into a large set of classes corresponding to their respective textual stimuli can be difficult, as some objects may share very similar meanings. EEG, as a standalone signal, might not capture subtle details across similar objects, but it may have a more abstract representation of them. Hence, we propose to study EEG-to-Text decoding in an episodic manner, with episodes sampled hierarchically. Firstly, we found that while non-episodic evaluation fails, the model performs better when evaluated episodically, indicating the presence of hierarchical representations in EEG. Secondly, we observed that coarse-grained tasks were easier than fine-grained tasks, but this was true only for binary classification. As we bring more classes to the task, both fine-grained and coarse-grained tasks become equally difficult. Another key finding is that models perform better when we attempt mapping signals to abstract classes (\eg, instead of considering BOY, GIRL, MAN, WOMAN as different classes, consider them as one class - PERSON). This suggests that EEG representations are sensitive to abstraction.

Additionally, by adopting an episodic approach, we created the largest episodic evaluation framework in the EEG domain, facilitating the ML community in developing efficient episodic learning algorithms for EEG decoding tasks. In this regard, we found that self-supervised pretrained models can outperform complex meta-learning techniques in EEG-to-Text decoding.


Our research highlights the importance of abstraction depth in the context of EEG-to-Text decoding. We hope this work will inspire further studies to enhance the efficacy of EEG-based applications across various fields.

\begin{acks}
    We would like to thank Atharv Nangare from IIT Gandhinagar for introducing us to the PEERS dataset and for the time he spent discussing its applications.
\end{acks}


\bibliographystyle{ACM-Reference-Format}
\bibliography{main}

\appendix

\begin{figure*}[]
  \vskip 0.2in
  \begin{center}
    \centerline{\includegraphics[width=0.9\linewidth]{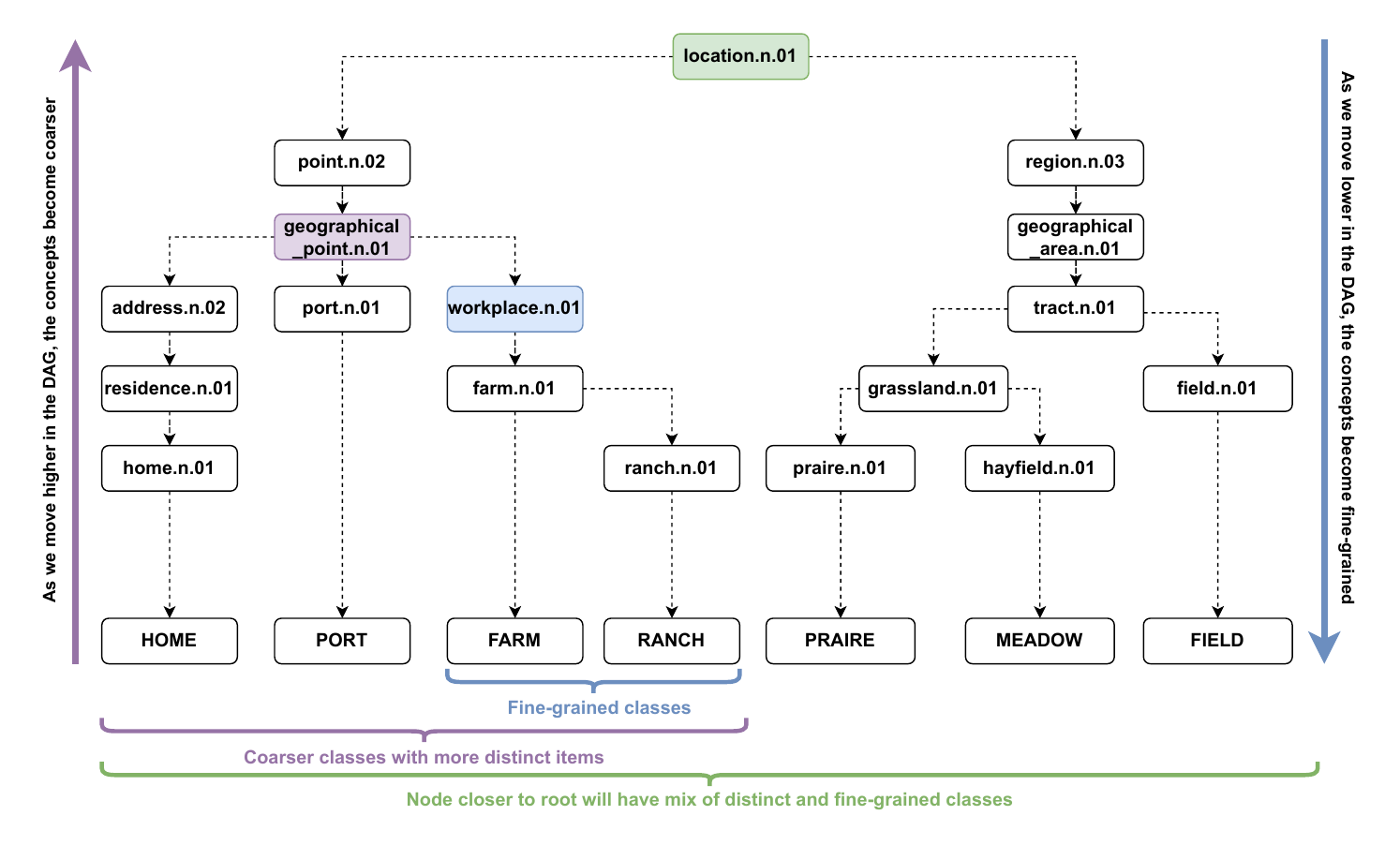}}
    \caption{
      Subgraph of the DAG for meta-train split showing leaf nodes sharing the same internal node having high similarity in meaning (\eg, FARM, RANCH). In the presence of multiple such instances, EEG might not capture representations for each of them, making non-episodic EEG-to-Text classification with all labels together a difficult task. 
    }
    \label{fig:sub_graph}
  \end{center}
\end{figure*}

\section{Dataset}
\label{sec:dataset}

The PEERS dataset was recorded via $5$ major recording experiments out of which we used the data of $4$ experiments whose description is provided below:

\textbf{Experiment $\mathbf{1}$ and $\mathbf{3}$:} Each session in the experiment consists of a series of 16 trials, each involving 16 visually presented words, followed by a recall task in which subjects recall as many words as they can. Each word is presented for $3$s, with an inter-stimulus interval (ISI) ranging from $0.8$s to $1.2$s. Each session consists of three kinds of trials - (i) \textit{$4$ No Task Trials}, where subject observes the presented words, (ii) \textit{$8$ Task Trials}, where in $4$ of the trials a subject decide if the object represented by the presented word fit in a shoebox (also known as size task) and in the remaining $4$ trials a subject decide if the object is living being (also known as animacy task), and (iii) \textit{$4$ Task-Shift Trials}, where both the size and animacy task exists within the same trial. Experiment $3$ differs from experiment $1$ in the recall task, where in experiment $3$, a subject also verbalizes the words they recall.

\textbf{Experiment $\mathbf{2}$:} This experiment introduces distractor intervals in addition to the word-based intervals in a trial. In the distractor interval, a math problem appeared on the screen in the form of $A+B+C=\mathit{?}$ where $A$, $B$, and $C$ are positive single-digit numbers.

\textbf{Experiment $\mathbf{4}$:} Experiment $4$ used a pool of $576$ words selected from the original pool. Each word is displayed to all $98$ subjects involved in this experiment. Each session had $24$ trials, with each trial presenting $24$ words individually for $1.6$s, with ISI ranging from $0.8$s to $1.2$s, followed by a distractor interval.

Experiments $1$, $2$, and $3$ of the PEERS dataset is also referred as (\textit{ltpFR}) and experiment $4$ as (\textit{ltpFR2}).

\section{Discarded Labels and Nodes in DAG}

\label{sec:app_dicarded_labels}

The word ``BLUEJAY" had no synonym in WordNet. Discarded words having too broad meaning are listed below, where the first value in the braces specifies the number of leaf nodes spanned by its parent, and the second value specifies the number of siblings:

\begin{itemize}
    \item CREATURE (121, 9)
    \item ANIMAL (121, 9)
    \item PERSON (278, 58)
    \item BEAST (121, 9)
    \item MAMMAL (56, 2)
\end{itemize}

The discarded internal nodes correspond to the following synonyms/concepts: `whole.n.02', `object.n.01', `physical\_entity.n.01'. The removal of these nodes also leads to the removal of $22$ more words (still retaining $\sim 98\%$ of total words in the PEERS Dataset).

\section{Hyperparameter Settings}

\begin{table}[]
\caption{Hyperparameter settings for experiments with non-episodic evaluation}
\label{tab:hp_non_episodic}
\begin{tabular}{@{}ll@{}}
\toprule
Hyperparamters & Settings \\ \midrule
Embedding dim  & 1024     \\
Dropout        & 0.05     \\
Learning rate  & 0.0003   \\
Epoch          & 50       \\
Weight Decay   & 0.001    \\
Optimizer      & AdamW    \\ \bottomrule
\end{tabular}
\end{table}

\begin{table}[]
\caption{Hyperparameter settings for experiments with episodic evaluation}
\label{tab:hp_episodic}
\resizebox{0.8\columnwidth}{!}{%
\begin{tabular}{@{}lll@{}}
\toprule
Learning Method                                     & Hyperparameters               & Settings         \\ \midrule
\multicolumn{1}{c}{\multirow{2}{*}{Common to both}} & Embedding Dim                 & 512              \\
\multicolumn{1}{c}{}                                & Dropout                       & 0.05             \\ \hline
\multicolumn{1}{c}{\multirow{4}{*}{Baseline}}       & Learning Rate                 & 0.0003           \\
\multicolumn{1}{c}{}                                & Epochs                        & 50               \\
\multicolumn{1}{c}{}                                & Batch Size                    & 128              \\
\multicolumn{1}{c}{}                                & Optimizer                     & AdamW            \\ \hline
\multicolumn{1}{c}{Meta-Learning}                   & Outer Learning Rate           & 0.0003           \\
                                                    & Inner Learning Rate           & 0.01             \\
                                                    & No. of inner updates          & 5                \\
                                                    & Inner Optimizer               & Gradient Descent \\
                                                    & Outer Optimizer               & Adam             \\
                                                    & Total Meta Steps              & 5351             \\
                                                    & Number of episodes in a Batch & 4                \\ \bottomrule
\end{tabular}%
}
\end{table}

In this section, we provide details related to the experiment setup. \Cref{tab:hp_non_episodic} and \cref{tab:hp_episodic} provide the hyper-parameter settings used for training the models via non-episodic and episodic learning, respectively.

\section{Words Under Moderately Broad Concepts}
\label{sec:span_set_lists}

This section lists the words under the span set of moderately broad concepts.

\textbf{matter.49:} SALAD, MAPLE, PUDDING, COFFEE, ORANGE, PASTA, BRANDY, GRAPE, PORK, PLASTER, PAPER, KLEENEX, BEAVER, BARLEY, LAVA, TREAT, GARBAGE, PIZZA, SUPPER, CANDY, OZONE, JELLO, HONEY, MEAT, DRUG, LUNCH, POISON, PASTRY, SODA, TROUT, QUAIL, SNACK, OINTMENT, SLIME, CUSTARD, CIGAR, JELLY, POWDER, SPONGE, SALT, ATOM, GLASS, CHAMPAGNE, COCKTAIL, KETCHUP, APPLE, BACON, RADISH, GREASE

\textbf{causal\_agent.72:}  VIKING, REBEL, SPHINX, VIRUS, CONVICT, FRIEND, MUMMY, PRINCESS, DIVER, PATIENT, LODGE, BRANDY, INFANT, BABY, WIFE, WASHER, PISTON, HIKER, SERVER, DENTIST, HUSBAND, HOSTESS, CAPTIVE, SISTER, DANCER, WOMAN, JUGGLER, LOVER, DRIVER, DRUG, ACTOR, AGENT, SIBLING, CATCHER, DINER, GIRL, PUPIL, LADY, INMATE, SPOUSE, WAITRESS, DAUGHTER, GANGSTER, ACTRESS, MARINE, TEACHER, OINTMENT, SERVANT, EXPERT, CASHIER, CIGAR, KEEPER, HOOD, CHEMIST, CHAUFFEUR, VAGRANT, POET, NOMAD, COCKTAIL, CHAMPAGNE, MAILMAN, PILOT, PRINCE, WORKER, PARTNER, WITNESS, BANKER, PREACHER, TYPIST, DONOR, OUTLAW, TOASTER

\textbf{person.62:} VIKING, REBEL, SPHINX, CONVICT, FRIEND, MUMMY, PRINCESS, DIVER, PATIENT, LODGE, INFANT, BABY, WIFE, WASHER, PISTON, HIKER, SERVER, DENTIST, HUSBAND, HOSTESS, CAPTIVE, SISTER, DANCER, WOMAN, JUGGLER, LOVER, ACTOR, SIBLING, CATCHER, DINER, GIRL, PUPIL, LADY, INMATE, SPOUSE, WAITRESS, DAUGHTER, GANGSTER, ACTRESS, MARINE, TEACHER, SERVANT, EXPERT, CASHIER, KEEPER, HOOD, CHEMIST, VAGRANT, POET, NOMAD, MAILMAN, PILOT, PRINCE, WORKER, PARTNER, WITNESS, BANKER, PREACHER, TYPIST, DONOR, OUTLAW, TOASTER



\appendix

\end{document}